\pdfoutput=1

\documentclass[11pt]{article}

\usepackage{EMNLP2023}

\usepackage{times}
\usepackage{latexsym}

\usepackage[T1]{fontenc}

\usepackage[utf8]{inputenc}

\usepackage{microtype}

\usepackage{inconsolata}

\usepackage{amsfonts}
\usepackage{amsmath}
\usepackage{arydshln}
\usepackage{bbding}
\usepackage{graphicx}
\usepackage{multirow}

\title{ACT-SQL: In-Context Learning for Text-to-SQL with Automatically-Generated Chain-of-Thought}

\author{Hanchong Zhang, Ruisheng Cao, Lu Chen\thanks{\ \ The corresponding author is Lu Chen.}, Hongshen Xu and Kai Yu \\
X-LANCE Lab, Department of Computer Science and Engineering \\
MoE Key Lab of Artificial Intelligence, SJTU AI Institute \\
Shanghai Jiao Tong University, Shanghai, China \\
{\tt \{zhanghanchong,chenlusz\}@sjtu.edu.cn}}

\begin{document}
\maketitle
\begin{abstract}
Recently Large Language Models (LLMs) have been proven to have strong abilities in various domains and tasks. We study the problem of prompt designing in the text-to-SQL task and attempt to improve the LLMs' reasoning ability when generating SQL queries. Besides the trivial few-shot in-context learning setting, we design our chain-of-thought (CoT) prompt with a similar method to schema linking. We provide a method named ACT-SQL\footnote{Our code is publicly available at \url{https://github.com/X-LANCE/text2sql-GPT}} to automatically generate auto-CoT exemplars and thus the whole process doesn't need manual labeling. Our approach is cost-saving since we only use the LLMs' API call once when generating one SQL query. Furthermore, we extend our in-context learning method to the multi-turn text-to-SQL task. The experiment results show that the LLMs' performance can benefit from our ACT-SQL approach. Our approach achieves SOTA performance on the Spider dev set among existing in-context learning approaches.
\end{abstract}

\section{Introduction}

The text-to-SQL task~\cite{zhongSeq2SQL2017, xu2017sqlnet} aims to translate the natural language question into the corresponding SQL query with the given database schema. It is the key technique to establish the natural language interface on relational databases, which can help common users access data from relational databases in a more convenient way.

Recent studies in text-to-SQL research have primarily centered on the development of semantic parsers within the framework of cross-domain analysis. In cross-domain text-to-SQL datasets such as Spider~\cite{yu-etal-2018-spider}, SParC~\cite{yu-etal-2019-sparc}, and CoSQL~\cite{yu-etal-2019-cosql}, the databases employed in the train set, dev set, and test set do not overlap. Prior research endeavors have focused on training specialized text-to-SQL models and optimizing their structural components to enhance overall performance. Notably, these efforts have yielded impressive model performances across various datasets. Nevertheless, the construction of such models necessitates a substantial number of high-quality training examples and entails significant time investments for finetuning. Moreover, these models often possess intricate structures, rendering their deployment challenging.

Recent research has provided empirical evidence establishing the substantial capabilities of Large Language Models (LLMs), such as GPT-3~\cite{brown2020language} and ChatGPT~\cite{ouyang2022training}, across a wide spectrum of domains and tasks. As the scale of LLMs continues to expand, scholarly investigations have revealed the presence of emergent abilities~\cite{wei2022emergent} exclusive to larger LLMs and absent in their smaller counterparts. Therefore, the latest studies employ LLMs in the context of the text-to-SQL task, utilizing the in-context learning method~\cite{brown2020language}. Owing to the impressive performance demonstrated by LLMs in zero-shot or few-shot prompting scenarios, the need for extensive finetuning using an abundance of training examples has been rendered unnecessary. Consequently, the integration of LLMs in the text-to-SQL process yields notable time and cost savings.

Nonetheless, contemporary in-context learning approaches for text-to-SQL encounter certain challenges. For instance, ~\citet{rajkumar2022evaluating}, in comparison to SOTA finetuned models, employ a simplistic prompt designing approach that yields relatively subpar performance. Similarly, ~\citet{pourreza2023dinsql} employs a convoluted workflow to generate the final SQL query, resulting in achieving SOTA performance on the test set of the Spider dataset. However, this approach proves time-consuming and resource-intensive, as it necessitates multiple API calls to LLMs during the query generation process. Moreover, the recent advancements in in-context learning methods for text-to-SQL have yet to be extended to multi-turn datasets, such as SParC, CoSQL, and DIR~\cite{li2023dir}.

Despite the proficiency of LLMs as zero-shot and few-shot learners, the mere superficial prompt design fails to fully activate their capabilities. To address this limitation, ~\citet{wei2023chainofthought} proposes a novel prompting technique called chain-of-thought (CoT). Through the CoT method, the prompt text encompasses a comprehensive thinking process that guides LLMs towards accurate deduction of answers. Notably, the CoT method mirrors the sequential nature of human reasoning, wherein intermediate answers are obtained before arriving at a final conclusion. Given the intricate nature of the text-to-SQL task, the CoT method proves highly suitable, as generating the SQL query entails complex reasoning processes. However, existing CoT methodologies necessitate extensive time investments in the selection of canonical examples and manual labeling. The text-to-SQL task lacks an automated approach for generating CoT sequences.

In this paper, we propose our in-context learning method for the text-to-SQL task with the automatically-generated CoT. First, under the zero-shot setting, we study the influence on LLMs' performance caused by the input format of the database schema. Second, under the few-shot setting, we provide a hybrid strategy to select exemplars and study the influence on LLMs' performance caused by the number of exemplars. Our experiment results show that the strategy is effective. Third, we present our approach named \textbf{ACT-SQL} to generate auto-CoT for the dataset training example consisting of the database schema, the question, and the corresponding SQL query. The experiment results show that the generated auto-CoT can indeed improve the LLMs' performance. The ACT-SQL achieves the SOTA performance on the Spider dev set among existing in-context learning methods. In addition, the ACT-SQL does not need to use extra LLMs' API calls, which means that our workflow is relatively fast and cheap. Finally, we apply our approach in multi-turn text-to-SQL datasets including SParC and CoSQL and achieve comparable accuracy scores with finetuned models. Our main contributions can be summarized as follows:

\begin{enumerate}
    \item We explore the influence on LLMs' performance under the text-to-SQL task with different prompting styles and few-shot exemplar selection strategies.
    \item We propose our approach named \textbf{ACT-SQL} to generate auto-CoT. The ACT-SQL achieves the SOTA performance on the Spider dev set among existing in-context learning methods. Furthermore, our automatic method is cost-saving and time-saving and does not need extra LLMs' API calls.
    \item We extend our method onto the multi-turn text-to-SQL task and achieve comparable performances with finetuned models on the SParC and CoSQL datasets.
\end{enumerate}

\section{Related Work}

\paragraph{Text-to-SQL models} Over the past several years, text-to-SQL researches mainly focus on building well-designed deep neural networks~\cite{chen-etal-2021-shadowgnn,cao2023heterogeneous}. RATSQL model~\cite{wang-etal-2020-rat} and LGESQL model~\cite{cao-etal-2021-lgesql} are AST-based approaches, where AST is the abbreviation of the abstract syntax tree. They encode the input and decode the AST of the SQL query with predefined grammar. AST-based approaches perform well but are generally complex to deploy.

PICARD~\cite{scholak-etal-2021-picard} is a sequence-to-sequence model. SQL is a formal language that follows strict grammar rules. Directly finetuning pretrained language models (PLMs) on text-to-SQL datasets would make PLMs likely to generate invalid SQL queries. The PICARD model rejects invalid tokens at each decoding step and constrains the generated results into a certain output space.

Although these specialized models have achieved excellent performances, there still exist some inevitable disadvantages. In order to train a text-to-SQL model, abundant high-quality training examples are needed. Constructing and labeling a large-scale text-to-SQL dataset is always not easy and would consume a lot of resources and time. Training and finetuning the model is also a hard project which costs many computing resources.

\paragraph{In-context learning for text-to-SQL} Since LLMs have shown amazing ability across various domains and have been applied in many academic and industrial fields, the latest researches begin to activate the LLMs' ability for the text-to-SQL task. ~\citet{rajkumar2022evaluating} uses the trivial zero-shot and few-shot learning setting and performs an empirical evaluation of text-to-SQL capabilities of LLMs including GPT-3~\cite{brown2020language} and Codex~\cite{chen2021evaluating}. They perform the zero-shot prompt learning on Spider~\cite{yu-etal-2018-spider}, a large-scale human-labeled cross-domain text-to-SQL dataset. Their work is relatively simple and the performance falls behind finetuned models.

~\citet{nan2023enhancing} mainly concentrates on the strategy of exemplars selection. Their work achieves good performance on several cross-domain datasets including Spider, Spider-Syn~\cite{gan-etal-2021-towards}, Spider-DK~\cite{gan-etal-2021-exploring} and Spider-Realistic~\cite{deng-etal-2021-structure}. However, their work requires an extra preliminary predictor to evaluate the SQL's difficulty level and needs to use LLMs' API call many times due to the majority vote method.

DIN-SQL~\cite{pourreza2023dinsql} provides a relatively complex approach. DIN-SQL consists of a complex workflow that decomposes the problem into several simpler sub-problems. With the LLM GPT-4, DIN-SQL has surpassed previous finetuned models and has achieved the best score on the Spider dataset. But DIN-SQL's workflow is obviously slow and expensive since it uses LLMs' API call many times to generate one SQL.

\section{Methodology}

With the in-context learning method, the SQL generation process can be formulated as
$$
S = \mathrm{LLM}(I, D, Q, \mathcal{E}).
$$
$I$ represents the instruction. $D$ represents the database schema. $Q$ represents the question. $\mathcal{E} = [(D_1, Q_1, P_1), \cdots, (D_n, Q_n, P_n)]$ is the list of exemplars where $P_i$ is the answer prompt which contains the correct SQL for the $i$-th exemplar. Thus the performance of LLMs is mainly influenced by the database prompt style, the exemplar selection strategy, and the exemplar prompt design.

In this section, we first describe the prompt styles of the database schema. Then we state our strategy of exemplar selection for the few-shot learning setting. Furthermore, we introduce our ACT-SQL approach, i.e. the automatically generated CoT method for constructing effective answer prompts. Finally, we extend our approach to the multi-turn text-to-SQL task.

\subsection{Database Prompt Style}
\label{sec:db_style}

Previous works have shown that given the database schema, strong LLMs (e.g. GPT models) can translate the relatively simple natural language question into the correct SQL query, though no exemplar is provided. Under the zero-shot setting, the LLMs merely take the database schema and the question as the input. Thus the input format of the database schema would mainly influence the LLMs' performance. Generally, we use five different database schema styles, which are shown in Appendix \ref{sec:app_db_style}:

\begin{enumerate}
    \item \texttt{Table(Column)} lists each table followed by its columns in each line. This style follows the official document provided by OpenAI\footnote{\url{https://platform.openai.com/examples/default-sql-translate}}.
    \item \texttt{Table(Column)(PF)} adds primary keys and foreign keys at the end of \texttt{Table(Column)}.
    \item \texttt{Create(NoPF)} describes all tables and columns with the ``create table'' statement in the SQL grammar. ``NoPF'' represents that no information on primary keys and foreign keys is added. Compared with \texttt{Table(Column)}, this input format contains information on column types (e.g. number and text) and is more similar to real SQL statements.
    \item \texttt{Create(EoC)} adds primary keys and foreign keys based on \texttt{Create(NoPF)}. ``EoC'' represents that they are added at the end of the corresponding column.
    \item \texttt{Create(EoT)} adds primary keys and foreign keys based on \texttt{Create(NoPF)}. ``EoT'' represents that they are added at the end of the table.
\end{enumerate}

Furthermore, database contents are concerned. Specifically $c$ example rows are appended to each table. Appendix \ref{sec:app_db_content} shows instances where $c = 3$.

\subsection{Exemplar Selection}

Given a few exemplars, LLMs can benefit and acquire tips from them and thus generate SQL queries with a more standard format and higher accuracy. Exemplar selection is an important work under the few-shot setting, which would influence the LLMs' performance a lot.

We select exemplars using a hybrid strategy. Specifically, we first of all select $n_s$ examples from the training dataset at random. These dataset examples are named static exemplars. They would be used in the context of every test case. As for each specific test case, we select $n_d$ extra examples from the training dataset. These dataset examples are named dynamic exemplars since they are selected according to some features of the current test case. Consequently, there are total $n_s + n_d$ exemplars for each test case.

In order to get dynamic exemplars that are more relevant to the current test case, we compare the natural language question of the current test case with all questions in the training dataset. We calculate the similarity scores with the suitable pretrained model and then select the top-$n_d$ training dataset examples. We believe that dynamic exemplars with more relevant questions would provide more effective information to the LLMs.

\subsection{Chain-of-Thought Prompt Design}

Under the few-shot learning setting, it has been proven that the LLMs' performance can benefit a lot from the chain-of-thought (CoT)~\cite{wei2023chainofthought} method. In the text-to-SQL task, only the database schema, the question, and the corresponding SQL query are provided in the prompt under the trivial few-shot learning setting. However, with the CoT method, the thought process of how to write the correct SQL query is added to the prompt. These prompting texts can help the LLMs think step by step when generating the complete SQL query and thus can activate the logical reasoning ability of the LLMs.

In previous works, some grammar-based text-to-SQL models utilize the graph encoding technique to jointly encode both the database schema and the question. Schema linking~\cite{bogin-etal-2019-representing,wang-etal-2020-rat,cao-etal-2021-lgesql} is a commonly used algorithm for building the input graph. If the question tokens exactly or partially match some schema item (i.e. table and column) names, then they are linked with the specific graph edge. It is obvious that the schema linking method can help the text-to-SQL models fetch the most relevant tables and columns among a great number of schema items based on the question.

\begin{figure}[htbp]
    \centering
    \includegraphics[width=0.49\textwidth]{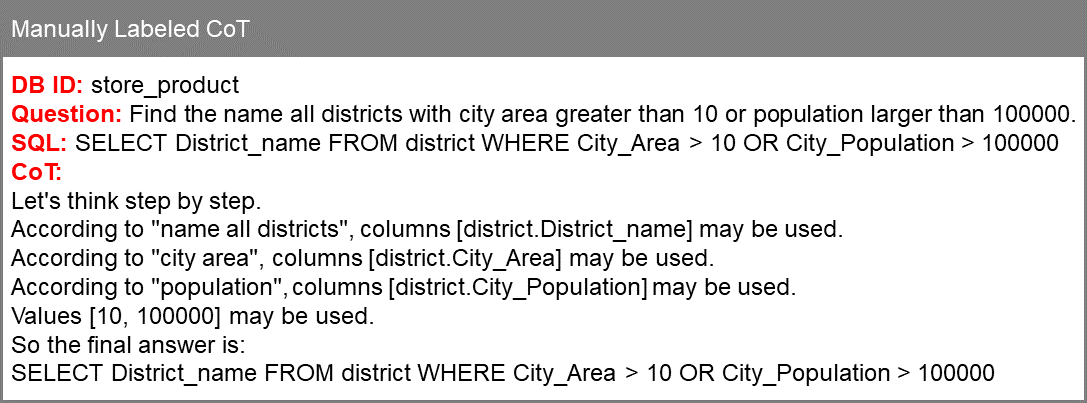}
    \caption{Manually labeled CoT for the dataset example.}
    \label{fig:manual_cot}
\end{figure}

We design our chain-of-thought prompt with a similar method to schema linking. Figure \ref{fig:manual_cot} shows an instance of the manually labeled CoT for the example from the train set of the Spider dataset~\cite{yu-etal-2018-spider}. As suggested in~\citet{kojima2023large}, the CoT prompt starts with ``Let's think step by step''. For each slice of the question sentence that may contain some information about the schema item, we add them into the CoT prompting text in the format shown in Figure \ref{fig:manual_cot}. Furthermore, the values mentioned in the question and the SQL query are also a concern. The final SQL query is appended at the end of the CoT prompt.

\paragraph{Auto-CoT} Although CoT prompts can be manually labeled, it costs a lot of time to find sufficient canonical and effective training dataset examples for CoT labeling. In addition, manually labeled CoT exemplars are fixed, which means that they are all static exemplars and dynamic exemplars are deficient. In order to deal with problems in the manual labeling process, we introduce an automatic method to generate auto-CoT prompts for every example in the training dataset.

Given the question $q = (q_1, q_2, \cdots, q_{|q|})$ and the SQL query $s$, the $q_i$ represents the $i$-th token in the question sentence. We define $q_{i, j} = (q_i, q_{i + 1}, \cdots, q_j)$ as a slice of the original question. We first enumerate each column $[tab].[col]$ appearing in the SQL query, where $[tab]$ represents the table name and $[col]$ represents the column name. For each column, we use the suitable pretrained model to compute the similarity scores between the current column and all the question sentence slices. The most relevant slice is
$$
\underset{q_{i, j}}{\arg \max} \ \ \mathrm{Sim}([tab].[col], q_{i, j}),
$$
where $\mathrm{Sim}$ is the similarity function. We link the column and its most relevant slice and add them to the auto-CoT prompt in the same format as the manual labeled CoT prompt. Note that during this process, we ignore the column appearing in the GROUP BY clause of the SQL query, since the GROUP BY column is commonly not mentioned directly in the question.

Secondly, we enumerate each table $[tab]$ appearing in the SQL query, where $[tab]$ represents the table name. In this process, we eliminate tables that have occurred in the columns, since those tables have been added into the auto CoT prompt. The left tables only appear in the FROM clause and indicate some extra information. For each table, we also compute all the similarity scores and find out the most relevant question slice, i.e.,
$$
\underset{q_{i, j}}{\arg \max} \ \  \mathrm{Sim}([tab], q_{i, j}).
$$
We link the table and its most relevant slice and add them to the auto-CoT.

\begin{figure}[htbp]
    \centering
    \includegraphics[width=0.49\textwidth]{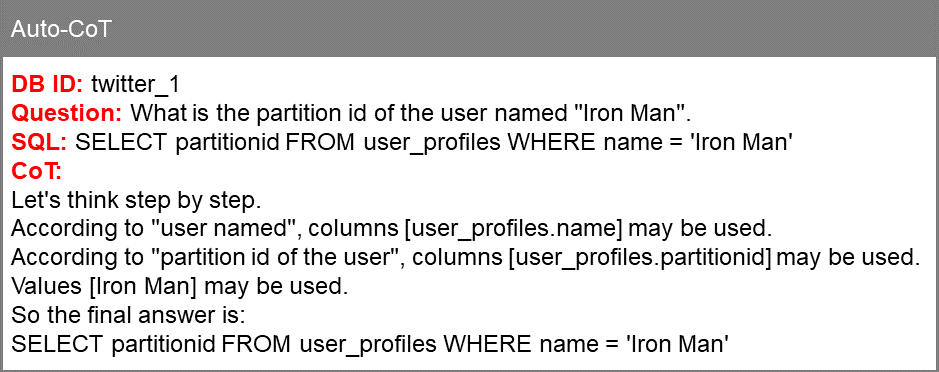}
    \caption{Auto-CoT for the dataset example.}
    \label{fig:auto_cot}
\end{figure}

Finally, we enumerate the values in the SQL query and then add them to the auto-CoT. Figure \ref{fig:auto_cot} shows an instance of the auto-generated CoT from the train set of the Spider dataset.

\subsection{Extension for Multi-turn Text-to-SQL}

The prompts described in the previous sections are designed for the single-turn text-to-SQL task. However, questions in the multi-turn text-to-SQL task are context-dependent and thus those prompts cannot be directly used. Moreover, the auto-CoT method is also disabled under the multi-turn setting, since the auto-CoT method finds information about schema linking based on the question slices. Under the multi-turn setting, this information may distribute into several context-dependent sentences.

In order to deal with the challenge of the multi-turn text-to-SQL task, we use LLMs to convert the multi-turn text-to-SQL task into the single-turn text-to-SQL task. Concretely, with the help of the LLMs, we can rewrite the question sentences and remove the context dependency among them. Thus each rewritten question and its corresponding SQL query turn into a new independent dataset example. We then directly apply the previous in-context learning method in the converted multi-turn text-to-SQL task.

The quality of the rewritten questions would influence the LLMs' performance a lot. It is necessary to manually label some rewriting exemplars in order to fix the format and improve the quality of the LLMs' outputs. For each multi-turn text-to-SQL dataset, we select 10 examples from the train set at random and manually label the rewritten results.

\section{Experiments}

\subsection{Experiment Setup}

\paragraph{Models} We mainly use the GPT-3.5-turbo model to evaluate our proposed approach. The GPT-3.5-turbo model is a low-cost LLM and is very large to have the emergent ability~\cite{wei2022emergent} for handling the text-to-SQL task. In addition, we use the GPT-4 model to evaluate our auto-CoT method on the Spider dataset~\cite{yu-etal-2018-spider}, since the GPT-4 model has a stronger reasoning ability but is much more expensive. We use the PLM \texttt{text2vec-base-chinese} to compute the similarity scores when selecting dynamic exemplars and generating auto-CoT prompts.

\paragraph{Hyperparameters} The temperature in LLMs' API is set to 0, i.e. the greedy decoding strategy is applied. The text-to-SQL tasks require the model to generate SQL queries with strict grammar rules. The LLMs are likely to generate invalid SQL queries or to write SQL queries that are not relevant to the given questions if the temperature is too high. The number of max tokens is set to 150 for the trivial in-context learning setting and 750 when using the CoT method.

\paragraph{Datasets} We mainly evaluate our proposed approach on Spider, a large-scale human-labeled cross-domain text-to-SQL dataset across 200 databases covering 138 domains. The Spider dataset contains 8,659 examples in the train set and 1,034 examples in the dev set. It also provides the evaluation script which divides SQL queries into four categories (i.e. easy, medium, hard, and extra) according to the difficulty level. The test set of Spider is not publicly available. We conduct the experiments on the dev set.

In addition, we also conduct the in-context learning experiments on Spider-Syn~\cite{gan-etal-2021-towards}, Spider-DK~\cite{gan-etal-2021-exploring} and Spider-Realistic~\cite{deng-etal-2021-structure}. Based on Spider, Spider-Syn replaces some schema-related tokens in the question with synonyms, which would make models unable to discover useful schema items with the simple string-matching method. Spider-DK defines five types of domain knowledge and modifies some examples by adding domain knowledge that reflects real-world question paraphrases. Spider-DK evaluates the models' generalization ability across domains when domain knowledge does not frequently appear in the train set. Spider-Realistic removes explicit mentions of column names to evaluate the model's ability to capture text-table alignment.

As for multi-turn text-to-SQL datasets, we conduct our experiments on SParC~\cite{yu-etal-2019-sparc} and CoSQL~\cite{yu-etal-2019-cosql}. SParC totally consists of 4,298 coherent question sequences including 12k+ individual questions and the corresponding SQL queries. CoSQL totally contains 10k+ annotated SQL queries. Each dialogue in CoSQL simulates a real-world scenario where the common user is exploring the database and the expert is retrieving answers with SQL.

\paragraph{Evaluation metrics} We use three commonly used evaluation metrics of the text-to-SQL task: exact match accuracy (EM), execution accuracy (EX), and test-suite accuracy (TS). The EM metric requires each component of the predicted SQL to be equivalent to the corresponding component of the gold SQL. Values in the SQL query are not concerned with the EM metric. The EX metric requires the execution result of the predicted SQL to be correct. Since there may exist different SQL queries that represent the same semantic, the EX metric is commonly more precise than the EM metric. The TS metric also evaluates the execution result but requires the result to be correct under multiple database instances per database schema\footnote{\url{https://github.com/taoyds/test-suite-sql-eval}}.

For multi-turn text-to-SQL datasets, we evaluate our approach with question match accuracy (QM) and interaction match accuracy (IM). The QM score is 1 if the predicted SQL query for the single question is correct. The IM score is 1 if all the predicted SQL queries in the interaction are correct.

\subsection{Zero-shot Results}





\begin{figure}[htbp]
    \centering
    \includegraphics[width=0.49\textwidth]{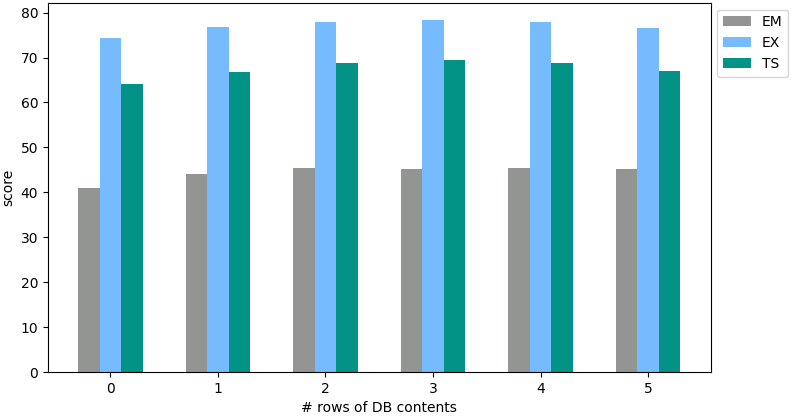}
    \caption{Zero-shot performances of GPT-3.5-turbo using \texttt{Table(Column)} DB style with different rows of DB contents on Spider dev set.}
    \label{fig:db_content}
\end{figure}

As discussed in Section \ref{sec:db_style}, the LLMs' performance is mainly influenced by the database prompt style and the rows of database contents under the zero-shot learning setting. We first conduct experiments for studying the influence on LLMs' performance caused by the rows of database contents. We fix the LLM as the GPT-3.5-turbo model and the database style as \texttt{Table(Column)} and only change the rows of database contents for each table in the prompt. Figure \ref{fig:db_content} shows the result on the Spider dev set. The LLM achieves the lowest score when no database content is provided. This indicates that database contents can provide useful tips for LLMs, especially when the testing case is sensitive to values in SQL where Table \ref{tab:db_content_case} shows two cases. In the first case, the 3 records from the database contain exactly one cell value ``France'' instead of ``French'' for the column ``singer.Citizenship''. Thus the LLM successfully predicts the correct value when these records are added to the prompt. In the second case, the database contents point out that ``Aberdeen'' is the city name so that the LLM can predict the correct SQL structure.

\begin{table}[htbp]
\centering
\resizebox{0.49\textwidth}{!}{
\begin{tabular}{p{\columnwidth}}
\hline

\hline
\textbf{Question:} {\it What are the names of the singers who are not French citizens?} \\
\textbf{DB content 0:} SELECT Name FROM singer WHERE Citizenship != '\textcolor[RGB]{176,23,31}{French}' \\
\textbf{DB content 3:} SELECT Name FROM singer WHERE Citizenship != '\textcolor[RGB]{0,112,192}{France}' \\
\hdashline
\textbf{Questions:} {\it Give the flight numbers of flights leaving from Aberdeen.} \\
\textbf{DB content 0:} SELECT FlightNo FROM flights WHERE \textcolor[RGB]{176,23,31}{SourceAirport = 'Aberdeen'} \\
\textbf{DB content 3:} SELECT FlightNo FROM flights WHERE \textcolor[RGB]{0,112,192}{SourceAirport IN (SELECT AirportCode FROM airports WHERE City = 'Aberdeen')} \\
\hline

\hline
\end{tabular}
}
\caption{Case study for different rows of DB contents.}
\label{tab:db_content_case}
\end{table}

The LLM gets the best score when the rows of database contents is set to 3. Too much database content in the prompt would not improve the LLMs' performance. Therefore, we always set the rows of database contents to 3 for the subsequent experiments.

\begin{table*}[htbp]
\centering
\begin{tabular}{c|c|c|ccc}
\hline

\hline
\multicolumn{3}{c|}{\textbf{Finetuned Model}} & \textbf{EM} & \textbf{EX} & \textbf{TS} \\
\hline

\hline
\multicolumn{3}{c|}{T5-3B+PICARD~\cite{scholak-etal-2021-picard}} & 75.5 & 79.3 & 71.9 \\
\multicolumn{3}{c|}{RASAT+PICARD~\cite{qi-etal-2022-rasat}} & 75.3 & 80.5 & - \\
\multicolumn{3}{c|}{N-best List Rerankers + PICARD~\cite{zeng2022nbest}} & 76.4 & 80.6 & - \\
\multicolumn{3}{c|}{Graphix-3B+PICARD~\cite{li2023graphixt5}} & 77.1 & 81.0 & - \\
\multicolumn{3}{c|}{RESDSQL-3B+NatSQL~\cite{li2023resdsql}} & \textbf{80.5} & \textbf{84.1} & - \\
\hline

\hline
\textbf{LLM} & \textbf{Approach} & \textbf{API per SQL} & \textbf{EM} & \textbf{EX} & \textbf{TS} \\
\hline

\hline
Codex Davinci & ~\citet{rajkumar2022evaluating} & 1 & - & 67.0 & 55.1 \\
Codex Davinci & ~\citet{chang2023prompt} & 1 & - & 76.8 & - \\
Codex Davinci & DIN-SQL~\cite{pourreza2023dinsql} & 4 & 57.2 & - & 69.9 \\
GPT-4 & DIN-SQL~\cite{pourreza2023dinsql} & 4 & 60.1 & 82.8 & 74.2 \\
GPT-3.5-turbo & \textbf{ACT-SQL (Ours)} & 1 & \textbf{62.7} & 80.4 & 71.4 \\
GPT-4 & \textbf{ACT-SQL (Ours)} & 1 & 61.7 & \textbf{82.9} & \textbf{74.5} \\
\hline

\hline
\end{tabular}
\caption{Performances of ACT-SQL and other previous works on Spider dev set.}
\label{tab:all_work}
\end{table*}

\begin{table}[htbp]
\centering
\begin{tabular}{c|ccc}
\hline

\hline
\textbf{DB Style} & \textbf{EM} & \textbf{EX} & \textbf{TS} \\
\hline

\hline
\texttt{Table(Column)} & 45.3 & 78.3 & \textbf{69.4} \\
\texttt{Table(Column)(PF)} & \textbf{45.4} & 79.0 & 69.1 \\
\texttt{Create(NoPF)} & 45.3 & 77.0 & 66.1 \\
\texttt{Create(EoC)} & 44.8 & \textbf{79.2} & 67.7 \\
\texttt{Create(EoT)} & 44.8 & 78.3 & 67.9 \\
\hline

\hline
\end{tabular}
\caption{Zero-shot performances of GPT-3.5-turbo with different DB styles on Spider dev set.}
\label{tab:db_style}
\end{table}

Table \ref{tab:db_style} shows the different performances of the GPT-3.5-turbo model when using different database styles. In general, \texttt{Table(Column)} and \texttt{Table(Column)(PF)} achieve higher scores than the other three database styles with the zero-shot learning setting because these two database styles follow the OpenAI's official document and may be more similar to LLMs' pretrained data. \texttt{Create(EoC)} and \texttt{Create(EoT)} performs better than \texttt{Create(NoPF)} in EX and TS metrics. This indicates that primary keys and foreign keys in the prompt can offer LLMs effective information.

\subsection{Few-shot Results}

Table \ref{tab:few_shot} shows all the few-shot experiment results on the Spider dev set, where different database styles and different numbers of static and dynamic exemplars are used. Compared with the zero-shot results, it is obvious that all the EM scores increase a lot. This is because SQL queries from the same dataset usually share similar grammar and structure and thus the exemplars from the Spider train set lead LLMs to output a similar SQL query.

Under the trivial few-shot learning setting, the TS scores also get improved by 1\%-3\% except for the \texttt{Table(Column)} database style. \texttt{Table(Column)} no longer performs better than \texttt{Table(Column)(PF)}, since LLMs' accuracy for predicting hard and extra hard SQL queries get increased with the few-shot exemplars and thus primary keys and foreign keys in the prompt become more important.

The experiment results prove that our ACT-SQL approach is effective. When the GPT-3.5-turbo model uses the ACT-SQL approach with the \texttt{Create(EoT)} database style, it achieves the best EM score of 62.7\% and the best TS score of 71.4\%. The best database style changes because LLMs can learn from exemplars. Table \ref{tab:auto_cot_case} shows the case study for the ACT-SQL method. With the trivial few-shot learning setting, there is a redundant column ``TV\_Channel.Hight\_definition\_TV'' appearing in the SELECT clause. When the ACT-SQL method is applied, the entire output generated by the LLM contains the complete thinking process which successfully does the schema linking. After clarifying all the tables and columns that may be used in SQL, the LLM eventually writes the correct SQL query without any redundant schema item.

Since the GPT-4 model is expensive, we use the GPT-4 model to evaluate our ACT-SQL approach only with the \texttt{Create(EoT)} database style and $n_s = n_d = 2$. Table \ref{tab:all_work} shows the performances of our ACT-SQL and other previous works using in-context learning with LLMs. The ACT-SQL approach uses the LLMs' API call only once for generating one SQL query and achieves the highest EM, EX, and TS scores among existing in-context learning approaches. ACT-SQL's performance is also comparable to finetuned models. Actually, finetuned models would get higher scores on the dev set than the test set, since these models are selected by the dev set performance. Instead, in-context learning methods would not suffer the performance gap between the dev set and the test set. Table \ref{tab:performance_gap} shows some previous works' performances on Spider dev set and test set. For finetuned approaches mentioned in the table, the performances drop from the dev set to the test set. On the contrary, for in-context learning approaches mentioned in the table, the performances increase from the dev set to the test set. After all, finetuned models are selected by the dev set performance, which would lead to the overfitting on the dev set and the performance dropping on the test set. For in-context learning approaches, the dev set and the test set are equal to the model. Performances between the dev set and the test set are only affected by the dataset feature.

\begin{table}[htbp]
\centering
\resizebox{0.49\textwidth}{!}{
\begin{tabular}{c|cc}
\hline

\hline
\textbf{Finetuned Approach} & \textbf{Dev} & \textbf{Test} \\
\hline

\hline
Graphix-3B+PICARD & 81.0 & 77.6 \\
RESDSQL-3B+NatSQL & 84.1 & 79.9 \\
\hline

\hline
\textbf{In-context Learning} & \textbf{Dev} & \textbf{Test} \\
\hline

\hline
C3~\cite{dong2023c3} & 81.8 & 82.3 \\
DIN-SQL~\cite{pourreza2023dinsql} & 82.8 & 85.3 \\
\hline

\hline
\end{tabular}
}
\caption{Performances of different previous approaches on Spider dev set and test set.}
\label{tab:performance_gap}
\end{table}

Table ~\ref{tab:spider_syn}, Table ~\ref{tab:spider_dk} and Table ~\ref{tab:spider_realistic} shows the GPT-3.5-turbo's performances on Spider-Syn, Spider-DK, and Spider-Realistic dev set. We use the \texttt{Create(EoT)} database style and set $n_s = n_d = 2$. The experiment results show that our approach is still comparable to finetuned models on Spider-Syn and Spider-Realistic datasets. On the Spider-DK dataset, our approach's EX score surpasses finetuned models. This is due to the wide range of domain knowledge stored in LLMs.

\begin{table}[htbp]
\centering
\begin{tabular}{c|ccc}
\hline

\hline
\textbf{Approach} & \textbf{EM} & \textbf{EX} & \textbf{TS} \\
\hline

\hline
Graphix-3B+PICARD & 66.9 & - & - \\
RESDSQL-3B+NatSQL & 69.1 & 76.9 & - \\
Few-shot (Ours) & 47.2 & 63.7 & 54.5 \\
ACT-SQL (Ours) & 51.5 & 67.9 & 59.3 \\
\hline

\hline
\end{tabular}
\caption{Performances of GPT-3.5-turbo and other previous works on Spider-Syn dev set.}
\label{tab:spider_syn}
\end{table}

\begin{table}[htbp]
\centering
\begin{tabular}{c|ccc}
\hline

\hline
\textbf{Approach} & \textbf{EM} & \textbf{EX} & \textbf{TS} \\
\hline

\hline
Graphix-3B+PICARD & 51.2 & - & - \\
RESDSQL-3B+NatSQL & 53.3 & 66.0 & - \\
Few-shot (Ours) & 50.7 & 68.8 & 61.7 \\
ACT-SQL (Ours) & 49.5 & 68.2 & 62.4 \\
\hline

\hline
\end{tabular}
\caption{Performances of GPT-3.5-turbo and other previous works on Spider-DK dev set.}
\label{tab:spider_dk}
\end{table}

\begin{table}[htbp]
\centering
\begin{tabular}{c|ccc}
\hline

\hline
\textbf{Approach} & \textbf{EM} & \textbf{EX} & \textbf{TS} \\
\hline

\hline
Graphix-3B+PICARD & 72.4 & - & - \\
RESDSQL-3B+NatSQL & 77.4 & 81.9 & - \\
Few-shot (Ours) & 52.4 & 76.4 & 62.0 \\
ACT-SQL (Ours) & 53.5 & 75.8 & 61.2 \\
\hline

\hline
\end{tabular}
\caption{Performances of GPT-3.5-turbo and other previous works on Spider-Realistic dev set.}
\label{tab:spider_realistic}
\end{table}

\subsection{Multi-turn Datasets Results}

\begin{table}[htbp]
\centering
\resizebox{0.49\textwidth}{!}{
\begin{tabular}{c|ccc|ccc}
\hline

\hline
\multirow{2}*{\textbf{Approach}} & \multicolumn{3}{c}{\textbf{QM}} & \multicolumn{3}{|c}{\textbf{IM}} \\
\cline{2-7}
~ & \textbf{EM} & \textbf{EX} & \textbf{TS} & \textbf{EM} & \textbf{EX} & \textbf{TS} \\
\hline

\hline
GAZP+BERT~\cite{zhong-etal-2020-grounded} & 48.9 & 47.8 & - & - & - & - \\
RASAT+PICARD~\cite{qi-etal-2022-rasat} & 67.7 & 73.3 & - & 49.1 & 54.0 & - \\
Few-shot (Ours) & 48.4 & 64.0 & 55.8 & 24.6 & 39.8 & 31.5 \\
ACT-SQL (Ours) & 51.0 & 63.8 & 56.9 & 24.4 & 38.9 & 29.6 \\
\hline

\hline
\end{tabular}
}
\caption{Performances of GPT-3.5-turbo and other previous works on SParC dev set.}
\label{tab:sparc}
\end{table}

\begin{table}[htbp]
\centering
\resizebox{0.49\textwidth}{!}{
\begin{tabular}{c|ccc|ccc}
\hline

\hline
\multirow{2}*{\textbf{Approach}} & \multicolumn{3}{c}{\textbf{QM}} & \multicolumn{3}{|c}{\textbf{IM}} \\
\cline{2-7}
~ & \textbf{EM} & \textbf{EX} & \textbf{TS} & \textbf{EM} & \textbf{EX} & \textbf{TS} \\
\hline

\hline
GAZP+BERT & 42.0 & 38.8 & - & - & - & - \\
RASAT+PICARD & 58.8 & 67.0 & - & 27.0 & 39.6 & - \\
Few-shot (Ours) & 45.4 & 63.9 & 55.5 & 14.7 & 29.4 & 22.9 \\
ACT-SQL (Ours) & 46.0 & 63.7 & 55.2 & 13.3 & 30.7 & 21.5 \\
\hline

\hline
\end{tabular}
}
\caption{Performances of GPT-3.5-turbo and other previous works on CoSQL dev set.}
\vspace{-5mm}
\label{tab:cosql}
\end{table}

Table \ref{tab:sparc} and Table \ref{tab:cosql} show the GPT-3.5-turbo's performances on two multi-turn text-to-SQL datasets, i.e. SParC and CoSQL. The database style is set to \texttt{Create(CoT)} and $n_d, n_s$ are set to 2 as before. The ACT-SQL approach is not that effective when applied to multi-turn datasets. We believe that our two-phase method causes bad performance. In the first phase, we use LLMs to rewrite questions in the interaction and convert the multi-turn dataset into the single-turn dataset. Sometimes the rewritten result's quality is bad, which influences the schema-linking process. Table \ref{tab:rewrite_case} shows two rewritten instances from the SParC dev set. In the first instance, the LLM correctly rewrites all sentences without missing any key information. However, in the second instance, the LLM does not remove the context dependency for the second sentence. This also leads to the error in the third sentence, where the keyword ``airline'' in the first sentence is missing. In general, our in-context learning method is comparable to finetuned models (GAZP + BERT) though there is still a big room for improvement. Improving LLMs' performance on this difficult task is a challenging future work. We just complete the initial exploration.

\begin{table}[htbp]
\centering
\resizebox{0.49\textwidth}{!}{
\begin{tabular}{p{\columnwidth}}
\hline

\hline
\textbf{Original questions:} \\
1. What are all the flights that leave from Aberdeen? \\
2. Of those, which land in Ashley? \\
3. How many are there? \\
\textbf{Rewritten questions:} \\
1. What are all the flights that depart from Aberdeen? \textcolor[RGB]{48,128,20}{\large\checkmark} \\
2. What are the flights that depart from Aberdeen and land in Ashley? \textcolor[RGB]{48,128,20}{\large\checkmark} \\
3. How many flights depart from Aberdeen and land in Ashley? \textcolor[RGB]{48,128,20}{\large\checkmark} \\
\hdashline
\textbf{Original questions:} \\
1. What are all the airlines? \\
2. Of these, which is Jetblue Airways? \\
3. What is the country corresponding it? \\
\textbf{Rewritten questions:} \\
1. What are all the airlines? \textcolor[RGB]{48,128,20}{\large\checkmark} \\
2. Which one of them is Jetblue Airways? \textcolor[RGB]{176,23,31}{\scriptsize\XSolidBold} \\
3. What is the country corresponding to Jetblue Airways? \textcolor[RGB]{176,23,31}{\scriptsize\XSolidBold} \\
\hline

\hline
\end{tabular}
}
\caption{Case study for rewritten questions from SParC dev set.}
\label{tab:rewrite_case}
\end{table}

\section{Conclusion}

LLMs have shown a strong ability in various domains with the in-context learning method. The latest studies have attempted to use LLMs to solve the text-to-SQL task. However, previous prompting approaches either perform worse than finetuned models or need to use LLMs' API call many times. We design the CoT prompt which can be automatically generated and propose our ACT-SQL approach. The ACT-SQL approach uses LLMs' API call only once to generate one SQL query. The experiment results prove that our approach achieves state-of-the-art performance on the Spider dev set among existing in-context learning approaches. Furthermore, we extend our approach to multi-turn text-to-SQL datasets.

\section*{Limitations}

There are some limitations in our work. First of all, we use a hybrid strategy for the exemplar selection. The numbers of static and dynamic exemplars are hyperparameters and still need manually determined. In addition, it is a relatively simple strategy that still needs improvement. Furthermore, our approach achieves relatively poor scores on some robustness variants of the Spider dataset and some multi-turn text-to-SQL datasets. Exploration of these datasets can be conducted in future work.

\section*{Acknowledgements}

We thank all the anonymous reviewers for their thoughtful comments. This work has been supported by the China NSFC Project (No.62106142 and No.62120106006), Shanghai Municipal Science and Technology Major Project (2021SHZDZX0102), and Startup Fund for Youngman Research at SJTU (SFYR at SJTU).

\bibliography{anthology,custom}
\bibliographystyle{acl_natbib}

\appendix
\onecolumn

\section{Detailed Experiment Results}

\begin{table*}[htbp]
\centering
\begin{tabular}{c|cc|ccc|ccc}
\hline

\hline
\multirow{2}*{\textbf{DB Style}} & \multirow{2}*{$n_s$} & \multirow{2}*{$n_d$} & \multicolumn{3}{c}{\textbf{Few-shot}} & \multicolumn{3}{|c}{\textbf{ACT-SQL}} \\
\cline{4-9}
~ & ~ & ~ & \textbf{EM} & \textbf{EX} & \textbf{TS} & \textbf{EM} & \textbf{EX} & \textbf{TS} \\
\hline

\hline
\multirow{3}*{\texttt{Table(Column)}} & 0 & 4 & 58.0 & 77.5 & 68.8 & 58.3 & 78.9 & 69.2 \\
~ & 2 & 2 & 59.7 & 77.9 & 68.7 & 60.7 & 78.7 & 69.4 \\
~ & 4 & 0 & 55.1 & 78.6 & 68.6 & 55.9 & 78.1 & 69.4 \\
\hline
\multirow{3}*{\texttt{Table(Column)(PF)}} & 0 & 4 & 59.3 & 79.9 & 70.6 & 60.8 & 79.8 & 70.2 \\
~ & 2 & 2 & 60.1 & 79.1 & 69.3 & 62.3 & 79.1 & 71.0 \\
~ & 4 & 0 & 54.8 & 80.0 & 70.3 & 57.0 & 77.9 & 69.6 \\
\hline
\multirow{3}*{\texttt{Create(NoPF)}} & 0 & 4 & 58.8 & 78.2 & 68.8 & 58.8 & 79.8 & 68.7 \\
~ & 2 & 2 & 60.0 & 79.2 & 68.9 & 60.6 & 79.1 & 69.1 \\
~ & 4 & 0 & 56.3 & 79.4 & 69.8 & 57.4 & 80.7 & 70.7 \\
\hline
\multirow{3}*{\texttt{Create(EoC)}} & 0 & 4 & 58.5 & 79.5 & 69.0 & 58.6 & 78.7 & 68.7 \\
~ & 2 & 2 & 60.1 & 79.4 & 68.7 & 61.9 & 79.0 & 71.1 \\
~ & 4 & 0 & 55.1 & 79.5 & 68.9 & 58.1 & \textbf{80.8} & 71.0 \\
\hline
\multirow{3}*{\texttt{Create(EoT)}} & 0 & 4 & 59.4 & 79.3 & 69.5 & 60.3 & 79.1 & 69.8 \\
~ & 2 & 2 & 61.4 & 79.4 & 69.0 & \textbf{62.7} & 80.4 & \textbf{71.4} \\
~ & 4 & 0 & 56.6 & 80.1 & 69.6 & 55.5 & 78.0 & 68.5 \\
\hline

\hline
\end{tabular}
\caption{Few-shot performances of GPT-3.5-turbo with different DB styles and different numbers of static and dynamic exemplars on Spider dev set.}
\label{tab:few_shot}
\end{table*}

\begin{table*}[htbp]
\centering
\begin{tabular}{cc|ccccc}
\hline

\hline
\textbf{LLM} & \textbf{Approach} & \textbf{Easy} & \textbf{Medium} & \textbf{Hard} & \textbf{Extra} & \textbf{All} \\
\hline

\hline
GPT-3.5-turbo & ACT-SQL (Ours) & 89.1 & 77.8 & 62.6 & 36.7 & 71.4 \\
GPT-4 & ACT-SQL (Ours) & 91.1 & 79.4 & 67.2 & 44.0 & 74.5 \\
\hline

\hline
\end{tabular}
\caption{Detailed performances of ACT-SQL on Spider dev set.}
\end{table*}

\section{Auto-CoT by LLM}

\begin{table}[htbp]
\centering
\resizebox{0.84\textwidth}{!}{
\begin{tabular}{p{\columnwidth}}
\hline

\hline
\textbf{Question:} {\it Find the package choice and series name of the TV channel that has high definition TV.} \\
\textbf{Few-shot:} SELECT series\_name, \textcolor[RGB]{176,23,31}{Hight\_definition\_TV}, Package\_Option FROM TV\_Channel WHERE Hight\_definition\_TV = 'yes' \\
\textbf{ACT-SQL:} SELECT package\_option , series\_name FROM TV\_Channel WHERE Hight\_definition\_TV = 'yes' \\
\textbf{Entire Output:} \\
Let's think step by step. \\
According to "TV channel that has high definition TV", columns [TV\_Channel.Hight\_definition\_TV] may be used. \\
According to "package choice and series name", columns [TV\_Channel.Package\_Option] and [TV\_Channel.series\_name] may be used. \\
So the final answer is: \\
SELECT package\_option , series\_name FROM TV\_Channel WHERE Hight\_definition\_TV = 'yes' \\
\hline

\hline
\end{tabular}
}
\caption{Case study for ACT-SQL. The entire output is totally generated by the LLM with the ACT-SQL approach.}
\label{tab:auto_cot_case}
\end{table}

\section{Prompts}

In this section, we list detailed prompts used in our experiments.

\subsection{Database Styles}
\label{sec:app_db_style}

\subsubsection{Table(Column)}

\# stadium(Stadium\_ID, Location, Name, Capacity, Highest, Lowest, Average) \\
\# singer(Singer\_ID, Name, Country, Song\_Name, Song\_release\_year, Age, Is\_male) \\
\# concert(concert\_ID, concert\_Name, Theme, Stadium\_ID, Year) \\
\# singer\_in\_concert(concert\_ID, Singer\_ID)

\subsubsection{Table(Column)(PF)}

\# stadium(Stadium\_ID, Location, Name, Capacity, Highest, Lowest, Average) \\
\# singer(Singer\_ID, Name, Country, Song\_Name, Song\_release\_year, Age, Is\_male) \\
\# concert(concert\_ID, concert\_Name, Theme, Stadium\_ID, Year) \\
\# singer\_in\_concert(concert\_ID, Singer\_ID) \\
\# primary keys = [stadium.Stadium\_ID, singer.Singer\_ID, concert.concert\_ID, singer\_in\_concert.concert\_ID] \\
\# foreign keys = [concert.Stadium\_ID = stadium.Stadium\_ID, singer\_in\_concert.Singer\_ID = singer.Singer\_ID, singer\_in\_concert.concert\_ID = concert.concert\_ID]

\subsubsection{Create(NoPF)}

create table stadium ( \\
\hspace*{2em} Stadium\_ID number, \\
\hspace*{2em} Location text, \\
\hspace*{2em} Name text, \\
\hspace*{2em} Capacity number, \\
\hspace*{2em} Highest number, \\
\hspace*{2em} Lowest number, \\
\hspace*{2em} Average number \\
) \\
create table singer ( \\
\hspace*{2em} Singer\_ID number, \\
\hspace*{2em} Name text, \\
\hspace*{2em} Country text, \\
\hspace*{2em} Song\_Name text, \\
\hspace*{2em} Song\_release\_year text, \\
\hspace*{2em} Age number, \\
\hspace*{2em} Is\_male others \\
) \\
create table concert ( \\
\hspace*{2em} concert\_ID number, \\
\hspace*{2em} concert\_Name text, \\
\hspace*{2em} Theme text, \\
\hspace*{2em} Stadium\_ID text, \\
\hspace*{2em} Year text \\
) \\
create table singer\_in\_concert ( \\
\hspace*{2em} concert\_ID number, \\
\hspace*{2em} Singer\_ID text \\
)

\subsubsection{Create(EoC)}

create table stadium ( \\
\hspace*{2em} Stadium\_ID number primary key, \\
\hspace*{2em} Location text, \\
\hspace*{2em} Name text, \\
\hspace*{2em} Capacity number, \\
\hspace*{2em} Highest number, \\
\hspace*{2em} Lowest number, \\
\hspace*{2em} Average number \\
) \\
create table singer ( \\
\hspace*{2em} Singer\_ID number primary key, \\
\hspace*{2em} Name text, \\
\hspace*{2em} Country text, \\
\hspace*{2em} Song\_Name text, \\
\hspace*{2em} Song\_release\_year text, \\
\hspace*{2em} Age number, \\
\hspace*{2em} Is\_male others \\
) \\
create table concert ( \\
\hspace*{2em} concert\_ID number primary key, \\
\hspace*{2em} concert\_Name text, \\
\hspace*{2em} Theme text, \\
\hspace*{2em} Stadium\_ID text references stadium(Stadium\_ID), \\
\hspace*{2em} Year text \\
) \\
create table singer\_in\_concert ( \\
\hspace*{2em} concert\_ID number primary key references concert(concert\_ID), \\
\hspace*{2em} Singer\_ID text references singer(Singer\_ID) \\
)

\subsubsection{Create(EoT)}

create table stadium ( \\
\hspace*{2em} Stadium\_ID number, \\
\hspace*{2em} Location text, \\
\hspace*{2em} Name text, \\
\hspace*{2em} Capacity number, \\
\hspace*{2em} Highest number, \\
\hspace*{2em} Lowest number, \\
\hspace*{2em} Average number, \\
\hspace*{2em} primary key (Stadium\_ID) \\
) \\
create table singer ( \\
\hspace*{2em} Singer\_ID number, \\
\hspace*{2em} Name text, \\
\hspace*{2em} Country text, \\
\hspace*{2em} Song\_Name text, \\
\hspace*{2em} Song\_release\_year text, \\
\hspace*{2em} Age number, \\
\hspace*{2em} Is\_male others, \\
\hspace*{2em} primary key (Singer\_ID) \\
) \\
create table concert ( \\
\hspace*{2em} concert\_ID number, \\
\hspace*{2em} concert\_Name text, \\
\hspace*{2em} Theme text, \\
\hspace*{2em} Stadium\_ID text, \\
\hspace*{2em} Year text, \\
\hspace*{2em} primary key (concert\_ID), \\
\hspace*{2em} foreign key (Stadium\_ID) references stadium(Stadium\_ID) \\
) \\
create table singer\_in\_concert ( \\
\hspace*{2em} concert\_ID number, \\
\hspace*{2em} Singer\_ID text, \\
\hspace*{2em} primary key (concert\_ID), \\
\hspace*{2em} foreign key (Singer\_ID) references singer(Singer\_ID), \\
\hspace*{2em} foreign key (concert\_ID) references concert(concert\_ID) \\
)

\subsection{Database Contents}
\label{sec:app_db_content}

We only use the \texttt{Table(Column)} and the \texttt{Create(EoT)} database styles in the following prompt examples. The other three database styles are similar. The rows of database contents is set to 3 in the following prompt examples.

\subsubsection{Table(Column)}

\# stadium(Stadium\_ID, Location, Name, Capacity, Highest, Lowest, Average) \\
/* \\
3 example rows from table stadium: \\
Stadium\_ID \hspace*{2em} Location \hspace*{2em} Name \hspace*{2em} Capacity \hspace*{2em} Highest \hspace*{2em} Lowest \hspace*{2em} Average \\
1 \hspace*{2em} Raith Rovers \hspace*{2em} Stark's Park \hspace*{2em} 10104 \hspace*{2em} 4812 \hspace*{2em} 1294 \hspace*{2em} 2106 \\
2 \hspace*{2em} Ayr United \hspace*{2em} Somerset Park \hspace*{2em} 11998 \hspace*{2em} 2363 \hspace*{2em} 1057 \hspace*{2em} 1477 \\
3 \hspace*{2em} East Fife \hspace*{2em} Bayview Stadium \hspace*{2em} 2000 \hspace*{2em} 1980 \hspace*{2em} 533 \hspace*{2em} 864 \\
**/ \\
\# singer(Singer\_ID, Name, Country, Song\_Name, Song\_release\_year, Age, Is\_male) \\
/* \\
3 example rows from table singer: \\
Singer\_ID \hspace*{2em} Name \hspace*{2em} Country \hspace*{2em} Song\_Name \hspace*{2em} Song\_release\_year \hspace*{2em} Age \hspace*{2em} Is\_male \\
1 \hspace*{2em} Joe Sharp \hspace*{2em} Netherlands \hspace*{2em} You \hspace*{2em} 1992 \hspace*{2em} 52 \hspace*{2em} F \\
2 \hspace*{2em} Timbaland \hspace*{2em} United States \hspace*{2em} Dangerous \hspace*{2em} 2008 \hspace*{2em} 32 \hspace*{2em} T \\
3 \hspace*{2em} Justin Brown \hspace*{2em} France \hspace*{2em} Hey Oh \hspace*{2em} 2013 \hspace*{2em} 29 \hspace*{2em} T \\
**/ \\
\# concert(concert\_ID, concert\_Name, Theme, Stadium\_ID, Year) \\
/* \\
3 example rows from table concert: \\
concert\_ID \hspace*{2em} concert\_Name \hspace*{2em} Theme \hspace*{2em} Stadium\_ID \hspace*{2em} Year \\
1 \hspace*{2em} Auditions \hspace*{2em} Free choice \hspace*{2em} 1 \hspace*{2em} 2014 \\
2 \hspace*{2em} Super bootcamp \hspace*{2em} Free choice 2 \hspace*{2em} 2 \hspace*{2em} 2014 \\
3 \hspace*{2em} Home Visits \hspace*{2em} Bleeding Love \hspace*{2em} 2 \hspace*{2em} 2015 \\
**/ \\
\# singer\_in\_concert(concert\_ID, Singer\_ID)
/* \\
3 example rows from table singer\_in\_concert: \\
concert\_ID \hspace*{2em} Singer\_ID \\
1 \hspace*{2em} 2 \\
1 \hspace*{2em} 3 \\
1 \hspace*{2em} 5 \\
**/

\subsubsection{Create(EoT)}

create table stadium ( \\
\hspace*{2em} Stadium\_ID number, \\
\hspace*{2em} Location text, \\
\hspace*{2em} Name text, \\
\hspace*{2em} Capacity number, \\
\hspace*{2em} Highest number, \\
\hspace*{2em} Lowest number, \\
\hspace*{2em} Average number, \\
\hspace*{2em} primary key (Stadium\_ID) \\
) \\
/* \\
3 example rows from table stadium: \\
Stadium\_ID \hspace*{2em} Location \hspace*{2em} Name \hspace*{2em} Capacity \hspace*{2em} Highest \hspace*{2em} Lowest \hspace*{2em} Average \\
1 \hspace*{2em} Raith Rovers \hspace*{2em} Stark's Park \hspace*{2em} 10104 \hspace*{2em} 4812 \hspace*{2em} 1294 \hspace*{2em} 2106 \\
2 \hspace*{2em} Ayr United \hspace*{2em} Somerset Park \hspace*{2em} 11998 \hspace*{2em} 2363 \hspace*{2em} 1057 \hspace*{2em} 1477 \\
3 \hspace*{2em} East Fife \hspace*{2em} Bayview Stadium \hspace*{2em} 2000 \hspace*{2em} 1980 \hspace*{2em} 533 \hspace*{2em} 864 \\
**/ \\
create table singer ( \\
\hspace*{2em} Singer\_ID number, \\
\hspace*{2em} Name text, \\
\hspace*{2em} Country text, \\
\hspace*{2em} Song\_Name text, \\
\hspace*{2em} Song\_release\_year text, \\
\hspace*{2em} Age number, \\
\hspace*{2em} Is\_male others, \\
\hspace*{2em} primary key (Singer\_ID) \\
) \\
/* \\
3 example rows from table singer: \\
Singer\_ID \hspace*{2em} Name \hspace*{2em} Country \hspace*{2em} Song\_Name \hspace*{2em} Song\_release\_year \hspace*{2em} Age \hspace*{2em} Is\_male \\
1 \hspace*{2em} Joe Sharp \hspace*{2em} Netherlands \hspace*{2em} You \hspace*{2em} 1992 \hspace*{2em} 52 \hspace*{2em} F \\
2 \hspace*{2em} Timbaland \hspace*{2em} United States \hspace*{2em} Dangerous \hspace*{2em} 2008 \hspace*{2em} 32 \hspace*{2em} T \\
3 \hspace*{2em} Justin Brown \hspace*{2em} France \hspace*{2em} Hey Oh \hspace*{2em} 2013 \hspace*{2em} 29 \hspace*{2em} T \\
**/ \\
create table concert ( \\
\hspace*{2em} concert\_ID number, \\
\hspace*{2em} concert\_Name text, \\
\hspace*{2em} Theme text, \\
\hspace*{2em} Stadium\_ID text, \\
\hspace*{2em} Year text, \\
\hspace*{2em} primary key (concert\_ID), \\
\hspace*{2em} foreign key (Stadium\_ID) references stadium(Stadium\_ID) \\
) \\
/* \\
3 example rows from table concert: \\
concert\_ID \hspace*{2em} concert\_Name \hspace*{2em} Theme \hspace*{2em} Stadium\_ID \hspace*{2em} Year \\
1 \hspace*{2em} Auditions \hspace*{2em} Free choice \hspace*{2em} 1 \hspace*{2em} 2014 \\
2 \hspace*{2em} Super bootcamp \hspace*{2em} Free choice 2 \hspace*{2em} 2 \hspace*{2em} 2014 \\
3 \hspace*{2em} Home Visits \hspace*{2em} Bleeding Love \hspace*{2em} 2 \hspace*{2em} 2015 \\
**/ \\
create table singer\_in\_concert ( \\
\hspace*{2em} concert\_ID number, \\
\hspace*{2em} Singer\_ID text, \\
\hspace*{2em} primary key (concert\_ID), \\
\hspace*{2em} foreign key (Singer\_ID) references singer(Singer\_ID), \\
\hspace*{2em} foreign key (concert\_ID) references concert(concert\_ID) \\
) \\
/* \\
3 example rows from table singer\_in\_concert: \\
concert\_ID \hspace*{2em} Singer\_ID \\
1 \hspace*{2em} 2 \\
1 \hspace*{2em} 3 \\
1 \hspace*{2em} 5 \\
**/

\subsection{In-context Learning Prompts}

We only use the \texttt{Create(EoT)} database styles in the following prompt examples. The other four database styles are similar. The rows of database contents is set to 3 in the following prompt examples. Under the few-shot setting, the first two shots are static exemplars and the last two shots are dynamic exemplars.

\subsubsection{Zero-shot}

role: system \\
content: \\
Given the database schema, you need to translate the question into the SQL query. \\
\\
role: user \\
content: \\
Database schema: \\
create table stadium ( \\
\hspace*{2em} Stadium\_ID number, \\
\hspace*{2em} Location text, \\
\hspace*{2em} Name text, \\
\hspace*{2em} Capacity number, \\
\hspace*{2em} Highest number, \\
\hspace*{2em} Lowest number, \\
\hspace*{2em} Average number, \\
\hspace*{2em} primary key (Stadium\_ID) \\
) \\
/* \\
3 example rows from table stadium: \\
Stadium\_ID \hspace*{2em} Location \hspace*{2em} Name \hspace*{2em} Capacity \hspace*{2em} Highest \hspace*{2em} Lowest \hspace*{2em} Average \\
1 \hspace*{2em} Raith Rovers \hspace*{2em} Stark's Park \hspace*{2em} 10104 \hspace*{2em} 4812 \hspace*{2em} 1294 \hspace*{2em} 2106 \\
2 \hspace*{2em} Ayr United \hspace*{2em} Somerset Park \hspace*{2em} 11998 \hspace*{2em} 2363 \hspace*{2em} 1057 \hspace*{2em} 1477 \\
3 \hspace*{2em} East Fife \hspace*{2em} Bayview Stadium \hspace*{2em} 2000 \hspace*{2em} 1980 \hspace*{2em} 533 \hspace*{2em} 864 \\
**/ \\
create table singer ( \\
\hspace*{2em} Singer\_ID number, \\
\hspace*{2em} Name text, \\
\hspace*{2em} Country text, \\
\hspace*{2em} Song\_Name text, \\
\hspace*{2em} Song\_release\_year text, \\
\hspace*{2em} Age number, \\
\hspace*{2em} Is\_male others, \\
\hspace*{2em} primary key (Singer\_ID) \\
) \\
/* \\
3 example rows from table singer: \\
Singer\_ID \hspace*{2em} Name \hspace*{2em} Country \hspace*{2em} Song\_Name \hspace*{2em} Song\_release\_year \hspace*{2em} Age \hspace*{2em} Is\_male \\
1 \hspace*{2em} Joe Sharp \hspace*{2em} Netherlands \hspace*{2em} You \hspace*{2em} 1992 \hspace*{2em} 52 \hspace*{2em} F \\
2 \hspace*{2em} Timbaland \hspace*{2em} United States \hspace*{2em} Dangerous \hspace*{2em} 2008 \hspace*{2em} 32 \hspace*{2em} T \\
3 \hspace*{2em} Justin Brown \hspace*{2em} France \hspace*{2em} Hey Oh \hspace*{2em} 2013 \hspace*{2em} 29 \hspace*{2em} T \\
**/ \\
create table concert ( \\
\hspace*{2em} concert\_ID number, \\
\hspace*{2em} concert\_Name text, \\
\hspace*{2em} Theme text, \\
\hspace*{2em} Stadium\_ID text, \\
\hspace*{2em} Year text, \\
\hspace*{2em} primary key (concert\_ID), \\
\hspace*{2em} foreign key (Stadium\_ID) references stadium(Stadium\_ID) \\
) \\
/* \\
3 example rows from table concert: \\
concert\_ID \hspace*{2em} concert\_Name \hspace*{2em} Theme \hspace*{2em} Stadium\_ID \hspace*{2em} Year \\
1 \hspace*{2em} Auditions \hspace*{2em} Free choice \hspace*{2em} 1 \hspace*{2em} 2014 \\
2 \hspace*{2em} Super bootcamp \hspace*{2em} Free choice 2 \hspace*{2em} 2 \hspace*{2em} 2014 \\
3 \hspace*{2em} Home Visits \hspace*{2em} Bleeding Love \hspace*{2em} 2 \hspace*{2em} 2015 \\
**/ \\
create table singer\_in\_concert ( \\
\hspace*{2em} concert\_ID number, \\
\hspace*{2em} Singer\_ID text, \\
\hspace*{2em} primary key (concert\_ID), \\
\hspace*{2em} foreign key (Singer\_ID) references singer(Singer\_ID), \\
\hspace*{2em} foreign key (concert\_ID) references concert(concert\_ID) \\
) \\
/* \\
3 example rows from table singer\_in\_concert: \\
concert\_ID \hspace*{2em} Singer\_ID \\
1 \hspace*{2em} 2 \\
1 \hspace*{2em} 3 \\
1 \hspace*{2em} 5 \\
**/ \\
Question: How many singers do we have?

\subsubsection{Few-shot}

role: system \\
content: \\
Given the database schema, you need to translate the question into the SQL query. \\
\\
role: user \\
content: \\
Database schema: \\
create table ACCOUNTS ( \\
\hspace*{2em} custid number, \\
\hspace*{2em} name text, \\
\hspace*{2em} primary key (custid) \\
) \\
/* \\
3 example rows from table ACCOUNTS: \\
custid \hspace*{2em} name \\
1 \hspace*{2em} Brown \\
2 \hspace*{2em} Wang \\
3 \hspace*{2em} O'mahony \\
**/ \\
create table SAVINGS ( \\
\hspace*{2em} custid number, \\
\hspace*{2em} balance number, \\
\hspace*{2em} primary key (custid), \\
\hspace*{2em} foreign key (custid) references ACCOUNTS(custid) \\
) \\
/* \\
3 example rows from table SAVINGS: \\
custid \hspace*{2em} balance \\
1 \hspace*{2em} 200000.0 \\
2 \hspace*{2em} 999999999.0 \\
3 \hspace*{2em} 230000.0 \\
**/ \\
create table CHECKING ( \\
\hspace*{2em} custid number, \\
\hspace*{2em} balance number, \\
\hspace*{2em} primary key (custid), \\
\hspace*{2em} foreign key (custid) references ACCOUNTS(custid) \\
) \\
/* \\
3 example rows from table CHECKING: \\
custid \hspace*{2em} balance \\
1 \hspace*{2em} 10000.0 \\
2 \hspace*{2em} 2000.0 \\
3 \hspace*{2em} 3000.0 \\
**/ \\
Question: Find the name and savings balance of the top 3 accounts with the highest saving balance sorted by savings balance in descending order. \\
\\
role: assistant \\
content: \\
SELECT T1.name ,  T2.balance FROM accounts AS T1 JOIN savings AS T2 ON T1.custid  =  T2.custid ORDER BY T2.balance DESC LIMIT 3 \\
\\
role: user \\
content: \\
Database schema: \\
create table flight ( \\
\hspace*{2em} flno number, \\
\hspace*{2em} origin text, \\
\hspace*{2em} destination text, \\
\hspace*{2em} distance number, \\
\hspace*{2em} departure\_date time, \\
\hspace*{2em} arrival\_date time, \\
\hspace*{2em} price number, \\
\hspace*{2em} aid number, \\
\hspace*{2em} primary key (flno), \\
\hspace*{2em} foreign key (aid) references aircraft(aid) \\
) \\
/* \\
3 example rows from table flight: \\
flno \hspace*{2em} origin \hspace*{2em} destination \hspace*{2em} distance \hspace*{2em} departure\_date \hspace*{2em} arrival\_date \hspace*{2em} price \hspace*{2em} aid \\
99 \hspace*{2em} Los Angeles \hspace*{2em} Washington D.C. \hspace*{2em} 2308 \hspace*{2em} 04/12/2005 09:30 \hspace*{2em} 04/12/2005 09:40 \hspace*{2em} 235.98 \hspace*{2em} 1 \\
13 \hspace*{2em} Los Angeles \hspace*{2em} Chicago \hspace*{2em} 1749 \hspace*{2em} 04/12/2005 08:45 \hspace*{2em} 04/12/2005 08:45 \hspace*{2em} 220.98 \hspace*{2em} 3 \\346 \hspace*{2em} Los Angeles \hspace*{2em} Dallas \hspace*{2em} 1251 \hspace*{2em} 04/12/2005 11:50 \hspace*{2em} 04/12/2005 07:05 \hspace*{2em} 182 \hspace*{2em} 2 \\
**/ \\
create table aircraft ( \\
\hspace*{2em} aid number, \\
\hspace*{2em} name text, \\
\hspace*{2em} distance number, \\
\hspace*{2em} primary key (aid) \\
) \\
/* \\
3 example rows from table aircraft: \\
aid \hspace*{2em} name \hspace*{2em} distance \\
1 \hspace*{2em} Boeing 747-400 \hspace*{2em} 8430 \\
2 \hspace*{2em} Boeing 737-800 \hspace*{2em} 3383 \\
3 \hspace*{2em} Airbus A340-300 \hspace*{2em} 7120 \\
**/ \\
create table employee ( \\
\hspace*{2em} eid number, \\
\hspace*{2em} name text, \\
\hspace*{2em} salary number, \\
\hspace*{2em} primary key (eid) \\
) \\
/* \\
3 example rows from table employee: \\
eid \hspace*{2em} name \hspace*{2em} salary \\
242518965 \hspace*{2em} James Smith \hspace*{2em} 120433 \\
141582651 \hspace*{2em} Mary Johnson \hspace*{2em} 178345 \\
11564812 \hspace*{2em} John Williams \hspace*{2em} 153972 \\
**/ \\
create table certificate ( \\
\hspace*{2em} eid number, \\
\hspace*{2em} aid number, \\
\hspace*{2em} primary key (eid), \\
\hspace*{2em} foreign key (aid) references aircraft(aid), \\
\hspace*{2em} foreign key (eid) references employee(eid) \\
) \\
/* \\
3 example rows from table certificate: \\
eid \hspace*{2em} aid \\
11564812 \hspace*{2em} 2 \\
11564812 \hspace*{2em} 10 \\
90873519 \hspace*{2em} 6 \\
**/ \\
Question: Which destination has least number of flights? \\
\\
role: assistant \\
content: \\
SELECT destination FROM Flight GROUP BY destination ORDER BY count(*) LIMIT 1 \\
\\
role: user \\
content: \\
Database schema: \\
create table Activity ( \\
\hspace*{2em} actid number, \\
\hspace*{2em} activity\_name text, \\
\hspace*{2em} primary key (actid) \\
) \\
/* \\
3 example rows from table Activity: \\
actid \hspace*{2em} activity\_name \\
770 \hspace*{2em} Mountain Climbing \\
771 \hspace*{2em} Canoeing \\
772 \hspace*{2em} Kayaking \\
**/ \\
create table Participates\_in ( \\
\hspace*{2em} stuid number, \\
\hspace*{2em} actid number, \\
\hspace*{2em} foreign key (actid) references Activity(actid), \\
\hspace*{2em} foreign key (stuid) references Student(StuID) \\
) \\
/* \\
3 example rows from table Participates\_in: \\
stuid \hspace*{2em} actid \\
1001 \hspace*{2em} 770 \\
1001 \hspace*{2em} 771 \\
1001 \hspace*{2em} 777 \\
**/ \\
create table Faculty\_Participates\_in ( \\
\hspace*{2em} FacID number, \\
\hspace*{2em} actid number, \\
\hspace*{2em} foreign key (actid) references Activity(actid), \\
\hspace*{2em} foreign key (FacID) references Faculty(FacID) \\
) \\
/* \\
3 example rows from table Faculty\_Participates\_in: \\
FacID \hspace*{2em} actid \\
1082 \hspace*{2em} 784 \\
1082 \hspace*{2em} 785 \\
1082 \hspace*{2em} 790 \\
**/ \\
create table Student ( \\
\hspace*{2em} StuID number, \\
\hspace*{2em} LName text, \\
\hspace*{2em} Fname text, \\
\hspace*{2em} Age number, \\
\hspace*{2em} Sex text, \\
\hspace*{2em} Major number, \\
\hspace*{2em} Advisor number, \\
\hspace*{2em} city\_code text, \\
\hspace*{2em} primary key (StuID) \\
) \\
/* \\
3 example rows from table Student: \\
StuID \hspace*{2em} LName \hspace*{2em} Fname \hspace*{2em} Age \hspace*{2em} Sex \hspace*{2em} Major \hspace*{2em} Advisor \hspace*{2em} city\_code \\
1001 \hspace*{2em} Smith \hspace*{2em} Linda \hspace*{2em} 18 \hspace*{2em} F \hspace*{2em} 600 \hspace*{2em} 1121 \hspace*{2em} BAL \\
1002 \hspace*{2em} Kim \hspace*{2em} Tracy \hspace*{2em} 19 \hspace*{2em} F \hspace*{2em} 600 \hspace*{2em} 7712 \hspace*{2em} HKG \\
1003 \hspace*{2em} Jones \hspace*{2em} Shiela \hspace*{2em} 21 \hspace*{2em} F \hspace*{2em} 600 \hspace*{2em} 7792 \hspace*{2em} WAS \\
**/ \\
create table Faculty ( \\
\hspace*{2em} FacID number, \\
\hspace*{2em} Lname text, \\
\hspace*{2em} Fname text, \\
\hspace*{2em} Rank text, \\
\hspace*{2em} Sex text, \\
\hspace*{2em} Phone number, \\
\hspace*{2em} Room text, \\
\hspace*{2em} Building text, \\
\hspace*{2em} primary key (FacID) \\
) \\
/* \\
3 example rows from table Faculty: \\
FacID \hspace*{2em} Lname \hspace*{2em} Fname \hspace*{2em} Rank \hspace*{2em} Sex \hspace*{2em} Phone \hspace*{2em} Room \hspace*{2em} Building \\
1082 \hspace*{2em} Giuliano \hspace*{2em} Mark \hspace*{2em} Instructor \hspace*{2em} M \hspace*{2em} 2424 \hspace*{2em} 224 \hspace*{2em} NEB \\
1121 \hspace*{2em} Goodrich \hspace*{2em} Michael \hspace*{2em} Professor \hspace*{2em} M \hspace*{2em} 3593 \hspace*{2em} 219 \hspace*{2em} NEB \\
1148 \hspace*{2em} Masson \hspace*{2em} Gerald \hspace*{2em} Professor \hspace*{2em} M \hspace*{2em} 3402 \hspace*{2em} 224B \hspace*{2em} NEB \\
**/ \\
Question: How many female Professors do we have? \\
\\
role: assistant \\
content: \\
SELECT count(*) FROM Faculty WHERE Sex  =  'F' AND Rank  =  "Professor" \\
\\
role: user \\
content: \\
Database schema: \\
create table region ( \\
\hspace*{2em} Region\_ID number, \\
\hspace*{2em} Region\_name text, \\
\hspace*{2em} Date text, \\
\hspace*{2em} Label text, \\
\hspace*{2em} Format text, \\
\hspace*{2em} Catalogue text, \\
\hspace*{2em} primary key (Region\_ID) \\
) \\
/* \\
3 example rows from table region: \\
Region\_ID \hspace*{2em} Region\_name \hspace*{2em} Date \hspace*{2em} Label \hspace*{2em} Format \hspace*{2em} Catalogue \\
1 \hspace*{2em} United Kingdom \hspace*{2em} 1 July 2002 \hspace*{2em} Parlophone \hspace*{2em} CD \hspace*{2em} 540 3622 \\
2 \hspace*{2em} United Kingdom \hspace*{2em} 1 July 2002 \hspace*{2em} Parlophone \hspace*{2em} 2× LP \hspace*{2em} 539 9821 \\
3 \hspace*{2em} Japan \hspace*{2em} 3 July 2002 \hspace*{2em} Toshiba-EMI \hspace*{2em} CD \hspace*{2em} TOCP-66045 \\
**/ \\
create table party ( \\
\hspace*{2em} Party\_ID number, \\
\hspace*{2em} Minister text, \\
\hspace*{2em} Took\_office text, \\
\hspace*{2em} Left\_office text, \\
\hspace*{2em} Region\_ID number, \\
\hspace*{2em} Party\_name text, \\
\hspace*{2em} primary key (Party\_ID), \\
\hspace*{2em} foreign key (Region\_ID) references region(Region\_ID) \\
) \\
/* \\
3 example rows from table party: \\
Party\_ID \hspace*{2em} Minister \hspace*{2em} Took\_office \hspace*{2em} Left\_office \hspace*{2em} Region\_ID \hspace*{2em} Party\_name \\
1 \hspace*{2em} Dr. Kwame Nkrumah (MP) \hspace*{2em} 1957 \hspace*{2em} 1958 \hspace*{2em} 1 \hspace*{2em} Convention Peoples Party \\
2 \hspace*{2em} Kojo Botsio (MP) \hspace*{2em} 1958 \hspace*{2em} 1959 \hspace*{2em} 2 \hspace*{2em} Progress Party \\
3 \hspace*{2em} Ebenezer Ako-Adjei (MP) \hspace*{2em} 1959 \hspace*{2em} 1960 \hspace*{2em} 3 \hspace*{2em} 3 \\
**/ \\
create table member ( \\
\hspace*{2em} Member\_ID number, \\
\hspace*{2em} Member\_Name text, \\
\hspace*{2em} Party\_ID text, \\
\hspace*{2em} In\_office text, \\
\hspace*{2em} primary key (Member\_ID), \\
\hspace*{2em} foreign key (Party\_ID) references party(Party\_ID) \\
) \\
/* \\
3 example rows from table member: \\
Member\_ID \hspace*{2em} Member\_Name \hspace*{2em} Party\_ID \hspace*{2em} In\_office \\
1 \hspace*{2em} Hon Tony Abbott \hspace*{2em} 3 \hspace*{2em} 1994–present \\
2 \hspace*{2em} Hon Dick Adams \hspace*{2em} 2 \hspace*{2em} 1993–2013 \\
3 \hspace*{2em} Anthony Albanese \hspace*{2em} 2 \hspace*{2em} 1996–present \\
**/ \\
create table party\_events ( \\
\hspace*{2em} Event\_ID number, \\
\hspace*{2em} Event\_Name text, \\
\hspace*{2em} Party\_ID number, \\
\hspace*{2em} Member\_in\_charge\_ID number, \\
\hspace*{2em} primary key (Event\_ID), \\
\hspace*{2em} foreign key (Member\_in\_charge\_ID) references member(Member\_ID), \\
\hspace*{2em} foreign key (Party\_ID) references party(Party\_ID) \\
) \\
/* \\
3 example rows from table party\_events: \\
Event\_ID \hspace*{2em} Event\_Name \hspace*{2em} Party\_ID \hspace*{2em} Member\_in\_charge\_ID \\
1 \hspace*{2em} Annaual Meeting \hspace*{2em} 1 \hspace*{2em} 4 \\
2 \hspace*{2em} Conference \hspace*{2em} 1 \hspace*{2em} 12 \\
3 \hspace*{2em} Annaual Meeting \hspace*{2em} 2 \hspace*{2em} 2 \\
**/ \\
Question: How many parties do we have? \\
\\
role: assistant \\
content: \\
SELECT count(DISTINCT party\_name) FROM party \\
\\
role: user \\
content: \\
Database schema: \\
create table stadium ( \\
\hspace*{2em} Stadium\_ID number, \\
\hspace*{2em} Location text, \\
\hspace*{2em} Name text, \\
\hspace*{2em} Capacity number, \\
\hspace*{2em} Highest number, \\
\hspace*{2em} Lowest number, \\
\hspace*{2em} Average number, \\
\hspace*{2em} primary key (Stadium\_ID) \\
) \\
/* \\
3 example rows from table stadium: \\
Stadium\_ID \hspace*{2em} Location \hspace*{2em} Name \hspace*{2em} Capacity \hspace*{2em} Highest \hspace*{2em} Lowest \hspace*{2em} Average \\
1 \hspace*{2em} Raith Rovers \hspace*{2em} Stark's Park \hspace*{2em} 10104 \hspace*{2em} 4812 \hspace*{2em} 1294 \hspace*{2em} 2106 \\
2 \hspace*{2em} Ayr United \hspace*{2em} Somerset Park \hspace*{2em} 11998 \hspace*{2em} 2363 \hspace*{2em} 1057 \hspace*{2em} 1477 \\
3 \hspace*{2em} East Fife \hspace*{2em} Bayview Stadium \hspace*{2em} 2000 \hspace*{2em} 1980 \hspace*{2em} 533 \hspace*{2em} 864 \\
**/ \\
create table singer ( \\
\hspace*{2em} Singer\_ID number, \\
\hspace*{2em} Name text, \\
\hspace*{2em} Country text, \\
\hspace*{2em} Song\_Name text, \\
\hspace*{2em} Song\_release\_year text, \\
\hspace*{2em} Age number, \\
\hspace*{2em} Is\_male others, \\
\hspace*{2em} primary key (Singer\_ID) \\
) \\
/* \\
3 example rows from table singer: \\
Singer\_ID \hspace*{2em} Name \hspace*{2em} Country \hspace*{2em} Song\_Name \hspace*{2em} Song\_release\_year \hspace*{2em} Age \hspace*{2em} Is\_male \\
1 \hspace*{2em} Joe Sharp \hspace*{2em} Netherlands \hspace*{2em} You \hspace*{2em} 1992 \hspace*{2em} 52 \hspace*{2em} F \\
2 \hspace*{2em} Timbaland \hspace*{2em} United States \hspace*{2em} Dangerous \hspace*{2em} 2008 \hspace*{2em} 32 \hspace*{2em} T \\
3 \hspace*{2em} Justin Brown \hspace*{2em} France \hspace*{2em} Hey Oh \hspace*{2em} 2013 \hspace*{2em} 29 \hspace*{2em} T \\
**/ \\
create table concert ( \\
\hspace*{2em} concert\_ID number, \\
\hspace*{2em} concert\_Name text, \\
\hspace*{2em} Theme text, \\
\hspace*{2em} Stadium\_ID text, \\
\hspace*{2em} Year text, \\
\hspace*{2em} primary key (concert\_ID), \\
\hspace*{2em} foreign key (Stadium\_ID) references stadium(Stadium\_ID) \\
) \\
/* \\
3 example rows from table concert: \\
concert\_ID \hspace*{2em} concert\_Name \hspace*{2em} Theme \hspace*{2em} Stadium\_ID \hspace*{2em} Year \\
1 \hspace*{2em} Auditions \hspace*{2em} Free choice \hspace*{2em} 1 \hspace*{2em} 2014 \\
2 \hspace*{2em} Super bootcamp \hspace*{2em} Free choice 2 \hspace*{2em} 2 \hspace*{2em} 2014 \\
3 \hspace*{2em} Home Visits \hspace*{2em} Bleeding Love \hspace*{2em} 2 \hspace*{2em} 2015 \\
**/ \\
create table singer\_in\_concert ( \\
\hspace*{2em} concert\_ID number, \\
\hspace*{2em} Singer\_ID text, \\
\hspace*{2em} primary key (concert\_ID), \\
\hspace*{2em} foreign key (Singer\_ID) references singer(Singer\_ID), \\
\hspace*{2em} foreign key (concert\_ID) references concert(concert\_ID) \\
) \\
/* \\
3 example rows from table singer\_in\_concert: \\
concert\_ID \hspace*{2em} Singer\_ID \\
1 \hspace*{2em} 2 \\
1 \hspace*{2em} 3 \\
1 \hspace*{2em} 5 \\
**/ \\
Question: How many singers do we have?

\subsubsection{Auto-CoT}

role: system \\
content: \\
Given the database schema, you need to translate the question into the SQL query. \\
\\
role: user \\
content: \\
Database schema: \\
create table ACCOUNTS ( \\
\hspace*{2em} custid number, \\
\hspace*{2em} name text, \\
\hspace*{2em} primary key (custid) \\
) \\
/* \\
3 example rows from table ACCOUNTS: \\
custid \hspace*{2em} name \\
1 \hspace*{2em} Brown \\
2 \hspace*{2em} Wang \\
3 \hspace*{2em} O'mahony \\
**/ \\
create table SAVINGS ( \\
\hspace*{2em} custid number, \\
\hspace*{2em} balance number, \\
\hspace*{2em} primary key (custid), \\
\hspace*{2em} foreign key (custid) references ACCOUNTS(custid) \\
) \\
/* \\
3 example rows from table SAVINGS: \\
custid \hspace*{2em} balance \\
1 \hspace*{2em} 200000.0 \\
2 \hspace*{2em} 999999999.0 \\
3 \hspace*{2em} 230000.0 \\
**/ \\
create table CHECKING ( \\
\hspace*{2em} custid number, \\
\hspace*{2em} balance number, \\
\hspace*{2em} primary key (custid), \\
\hspace*{2em} foreign key (custid) references ACCOUNTS(custid) \\
) \\
/* \\
3 example rows from table CHECKING: \\
custid \hspace*{2em} balance \\
1 \hspace*{2em} 10000.0 \\
2 \hspace*{2em} 2000.0 \\
3 \hspace*{2em} 3000.0 \\
**/ \\
Question: Find the name and savings balance of the top 3 accounts with the highest saving balance sorted by savings balance in descending order. \\
\\
role: assistant \\
content: \\
Let's think step by step. \\
According to "savings balance", columns [SAVINGS.balance] may be used. \\
According to "accounts", columns [ACCOUNTS.name] may be used. \\
Values [3] may be used. \\
So the final answer is: \\
SELECT T1.name , T2.balance FROM accounts AS T1 JOIN savings AS T2 ON T1.custid = T2.custid ORDER BY T2.balance DESC LIMIT 3 \\
\\
role: user \\
content: \\
Database schema: \\
create table flight ( \\
\hspace*{2em} flno number, \\
\hspace*{2em} origin text, \\
\hspace*{2em} destination text, \\
\hspace*{2em} distance number, \\
\hspace*{2em} departure\_date time, \\
\hspace*{2em} arrival\_date time, \\
\hspace*{2em} price number, \\
\hspace*{2em} aid number, \\
\hspace*{2em} primary key (flno), \\
\hspace*{2em} foreign key (aid) references aircraft(aid) \\
) \\
/* \\
3 example rows from table flight: \\
flno \hspace*{2em} origin \hspace*{2em} destination \hspace*{2em} distance \hspace*{2em} departure\_date \hspace*{2em} arrival\_date \hspace*{2em} price \hspace*{2em} aid \\
99 \hspace*{2em} Los Angeles \hspace*{2em} Washington D.C. \hspace*{2em} 2308 \hspace*{2em} 04/12/2005 09:30 \hspace*{2em} 04/12/2005 09:40 \hspace*{2em} 235.98 \hspace*{2em} 1 \\
13 \hspace*{2em} Los Angeles \hspace*{2em} Chicago \hspace*{2em} 1749 \hspace*{2em} 04/12/2005 08:45 \hspace*{2em} 04/12/2005 08:45 \hspace*{2em} 220.98 \hspace*{2em} 3 \\346 \hspace*{2em} Los Angeles \hspace*{2em} Dallas \hspace*{2em} 1251 \hspace*{2em} 04/12/2005 11:50 \hspace*{2em} 04/12/2005 07:05 \hspace*{2em} 182 \hspace*{2em} 2 \\
**/ \\
create table aircraft ( \\
\hspace*{2em} aid number, \\
\hspace*{2em} name text, \\
\hspace*{2em} distance number, \\
\hspace*{2em} primary key (aid) \\
) \\
/* \\
3 example rows from table aircraft: \\
aid \hspace*{2em} name \hspace*{2em} distance \\
1 \hspace*{2em} Boeing 747-400 \hspace*{2em} 8430 \\
2 \hspace*{2em} Boeing 737-800 \hspace*{2em} 3383 \\
3 \hspace*{2em} Airbus A340-300 \hspace*{2em} 7120 \\
**/ \\
create table employee ( \\
\hspace*{2em} eid number, \\
\hspace*{2em} name text, \\
\hspace*{2em} salary number, \\
\hspace*{2em} primary key (eid) \\
) \\
/* \\
3 example rows from table employee: \\
eid \hspace*{2em} name \hspace*{2em} salary \\
242518965 \hspace*{2em} James Smith \hspace*{2em} 120433 \\
141582651 \hspace*{2em} Mary Johnson \hspace*{2em} 178345 \\
11564812 \hspace*{2em} John Williams \hspace*{2em} 153972 \\
**/ \\
create table certificate ( \\
\hspace*{2em} eid number, \\
\hspace*{2em} aid number, \\
\hspace*{2em} primary key (eid), \\
\hspace*{2em} foreign key (aid) references aircraft(aid), \\
\hspace*{2em} foreign key (eid) references employee(eid) \\
) \\
/* \\
3 example rows from table certificate: \\
eid \hspace*{2em} aid \\
11564812 \hspace*{2em} 2 \\
11564812 \hspace*{2em} 10 \\
90873519 \hspace*{2em} 6 \\
**/ \\
Question: Which destination has least number of flights? \\
\\
role: assistant \\
content: \\
Let's think step by step. \\
According to "flights", columns [flight.destination] may be used. \\
Values [1] may be used. \\
So the final answer is: \\
SELECT destination FROM Flight GROUP BY destination ORDER BY count(*) LIMIT 1 \\
\\
role: user \\
content: \\
Database schema: \\
create table Student ( \\
\hspace*{2em} StuID number, \\
\hspace*{2em} LName text, \\
\hspace*{2em} Fname text, \\
\hspace*{2em} Age number, \\
\hspace*{2em} Sex text, \\
\hspace*{2em} Major number, \\
\hspace*{2em} Advisor number, \\
\hspace*{2em} city\_code text, \\
\hspace*{2em} primary key (StuID) \\
) \\
/* \\
3 example rows from table Student: \\
StuID \hspace*{2em} LName \hspace*{2em} Fname \hspace*{2em} Age \hspace*{2em} Sex \hspace*{2em} Major \hspace*{2em} Advisor \hspace*{2em} city\_code \\
1001 \hspace*{2em} Smith \hspace*{2em} Linda \hspace*{2em} 18 \hspace*{2em} F \hspace*{2em} 600 \hspace*{2em} 1121 \hspace*{2em} BAL \\
1002 \hspace*{2em} Kim \hspace*{2em} Tracy \hspace*{2em} 19 \hspace*{2em} F \hspace*{2em} 600 \hspace*{2em} 7712 \hspace*{2em} HKG \\
1003 \hspace*{2em} Jones \hspace*{2em} Shiela \hspace*{2em} 21 \hspace*{2em} F \hspace*{2em} 600 \hspace*{2em} 7792 \hspace*{2em} WAS \\
**/ \\
create table Dorm ( \\
\hspace*{2em} dormid number, \\
\hspace*{2em} dorm\_name text, \\
\hspace*{2em} student\_capacity number, \\
\hspace*{2em} gender text \\
) \\
/* \\
3 example rows from table Dorm: \\
dormid \hspace*{2em} dorm\_name \hspace*{2em} student\_capacity \hspace*{2em} gender \\
100 \hspace*{2em} Smith Hall \hspace*{2em} 85 \hspace*{2em} X \\
110 \hspace*{2em} Bud Jones Hall \hspace*{2em} 116 \hspace*{2em} M \\
140 \hspace*{2em} Fawlty Towers \hspace*{2em} 355 \hspace*{2em} X \\
**/ \\
create table Dorm\_amenity ( \\
\hspace*{2em} amenid number, \\
\hspace*{2em} amenity\_name text \\
) \\
/* \\
3 example rows from table Dorm\_amenity: \\
amenid \hspace*{2em} amenity\_name \\
900 \hspace*{2em} TV Lounge \\
901 \hspace*{2em} Study Room \\
902 \hspace*{2em} Pub in Basement \\
**/ \\
create table Has\_amenity ( \\
\hspace*{2em} dormid number, \\
\hspace*{2em} amenid number, \\
\hspace*{2em} foreign key (amenid) references Dorm\_amenity(amenid), \\
\hspace*{2em} foreign key (dormid) references Dorm(dormid) \\
) \\
/* \\
3 example rows from table Has\_amenity: \\
dormid \hspace*{2em} amenid \\
109 \hspace*{2em} 900 \\
109 \hspace*{2em} 901 \\
109 \hspace*{2em} 903 \\
**/ \\
create table Lives\_in ( \\
\hspace*{2em} stuid number, \\
\hspace*{2em} dormid number, \\
\hspace*{2em} room\_number number, \\
\hspace*{2em} foreign key (dormid) references Dorm(dormid), \\
\hspace*{2em} foreign key (stuid) references Student(StuID) \\
) \\
/* \\
3 example rows from table Lives\_in: \\
stuid \hspace*{2em} dormid \hspace*{2em} room\_number \\
1001 \hspace*{2em} 109 \hspace*{2em} 105 \\
1002 \hspace*{2em} 100 \hspace*{2em} 112 \\
1003 \hspace*{2em} 100 \hspace*{2em} 124 \\
**/ \\
Question: How many dorms have amenities? \\
\\
role: assistant \\
content: \\
Let's think step by step. \\
According to "dorms have amenities ?", columns [Has\_amenity.dormid] may be used. \\
So the final answer is: \\
SELECT count(DISTINCT dormid) FROM has\_amenity \\
\\
role: user \\
content: \\
Database schema: \\
create table Student ( \\
\hspace*{2em} StuID number, \\
\hspace*{2em} LName text, \\
\hspace*{2em} Fname text, \\
\hspace*{2em} Age number, \\
\hspace*{2em} Sex text, \\
\hspace*{2em} Major number, \\
\hspace*{2em} Advisor number, \\
\hspace*{2em} city\_code text, \\
\hspace*{2em} primary key (StuID) \\
) \\
/* \\
3 example rows from table Student: \\
StuID \hspace*{2em} LName \hspace*{2em} Fname \hspace*{2em} Age \hspace*{2em} Sex \hspace*{2em} Major \hspace*{2em} Advisor \hspace*{2em} city\_code \\
1001 \hspace*{2em} Smith \hspace*{2em} Linda \hspace*{2em} 18 \hspace*{2em} F \hspace*{2em} 600 \hspace*{2em} 1121 \hspace*{2em} BAL \\
1002 \hspace*{2em} Kim \hspace*{2em} Tracy \hspace*{2em} 19 \hspace*{2em} F \hspace*{2em} 600 \hspace*{2em} 7712 \hspace*{2em} HKG \\
1003 \hspace*{2em} Jones \hspace*{2em} Shiela \hspace*{2em} 21 \hspace*{2em} F \hspace*{2em} 600 \hspace*{2em} 7792 \hspace*{2em} WAS \\
**/ \\
create table Dorm ( \\
\hspace*{2em} dormid number, \\
\hspace*{2em} dorm\_name text, \\
\hspace*{2em} student\_capacity number, \\
\hspace*{2em} gender text \\
) \\
/* \\
3 example rows from table Dorm: \\
dormid \hspace*{2em} dorm\_name \hspace*{2em} student\_capacity \hspace*{2em} gender \\
100 \hspace*{2em} Smith Hall \hspace*{2em} 85 \hspace*{2em} X \\
110 \hspace*{2em} Bud Jones Hall \hspace*{2em} 116 \hspace*{2em} M \\
140 \hspace*{2em} Fawlty Towers \hspace*{2em} 355 \hspace*{2em} X \\
**/ \\
create table Dorm\_amenity ( \\
\hspace*{2em} amenid number, \\
\hspace*{2em} amenity\_name text \\
) \\
/* \\
3 example rows from table Dorm\_amenity: \\
amenid \hspace*{2em} amenity\_name \\
900 \hspace*{2em} TV Lounge \\
901 \hspace*{2em} Study Room \\
902 \hspace*{2em} Pub in Basement \\
**/ \\
create table Has\_amenity ( \\
\hspace*{2em} dormid number, \\
\hspace*{2em} amenid number, \\
\hspace*{2em} foreign key (amenid) references Dorm\_amenity(amenid), \\
\hspace*{2em} foreign key (dormid) references Dorm(dormid) \\
) \\
/* \\
3 example rows from table Has\_amenity: \\
dormid \hspace*{2em} amenid \\
109 \hspace*{2em} 900 \\
109 \hspace*{2em} 901 \\
109 \hspace*{2em} 903 \\
**/ \\
create table Lives\_in ( \\
\hspace*{2em} stuid number, \\
\hspace*{2em} dormid number, \\
\hspace*{2em} room\_number number, \\
\hspace*{2em} foreign key (dormid) references Dorm(dormid), \\
\hspace*{2em} foreign key (stuid) references Student(StuID) \\
) \\
/* \\
3 example rows from table Lives\_in: \\
stuid \hspace*{2em} dormid \hspace*{2em} room\_number \\
1001 \hspace*{2em} 109 \hspace*{2em} 105 \\
1002 \hspace*{2em} 100 \hspace*{2em} 112 \\
1003 \hspace*{2em} 100 \hspace*{2em} 124 \\
**/ \\
Question: How many diffrent dorm amenities are there? \\
\\
role: assistant \\
content: \\
Let's think step by step. \\
According to "dorm amenities are there ?", tables [Dorm\_amenity] may be used. \\
So the final answer is: \\
SELECT count(*) FROM dorm\_amenity \\
\\
role: user \\
content: \\
Database schema: \\
create table stadium ( \\
\hspace*{2em} Stadium\_ID number, \\
\hspace*{2em} Location text, \\
\hspace*{2em} Name text, \\
\hspace*{2em} Capacity number, \\
\hspace*{2em} Highest number, \\
\hspace*{2em} Lowest number, \\
\hspace*{2em} Average number, \\
\hspace*{2em} primary key (Stadium\_ID) \\
) \\
/* \\
3 example rows from table stadium: \\
Stadium\_ID \hspace*{2em} Location \hspace*{2em} Name \hspace*{2em} Capacity \hspace*{2em} Highest \hspace*{2em} Lowest \hspace*{2em} Average \\
1 \hspace*{2em} Raith Rovers \hspace*{2em} Stark's Park \hspace*{2em} 10104 \hspace*{2em} 4812 \hspace*{2em} 1294 \hspace*{2em} 2106 \\
2 \hspace*{2em} Ayr United \hspace*{2em} Somerset Park \hspace*{2em} 11998 \hspace*{2em} 2363 \hspace*{2em} 1057 \hspace*{2em} 1477 \\
3 \hspace*{2em} East Fife \hspace*{2em} Bayview Stadium \hspace*{2em} 2000 \hspace*{2em} 1980 \hspace*{2em} 533 \hspace*{2em} 864 \\
**/ \\
create table singer ( \\
\hspace*{2em} Singer\_ID number, \\
\hspace*{2em} Name text, \\
\hspace*{2em} Country text, \\
\hspace*{2em} Song\_Name text, \\
\hspace*{2em} Song\_release\_year text, \\
\hspace*{2em} Age number, \\
\hspace*{2em} Is\_male others, \\
\hspace*{2em} primary key (Singer\_ID) \\
) \\
/* \\
3 example rows from table singer: \\
Singer\_ID \hspace*{2em} Name \hspace*{2em} Country \hspace*{2em} Song\_Name \hspace*{2em} Song\_release\_year \hspace*{2em} Age \hspace*{2em} Is\_male \\
1 \hspace*{2em} Joe Sharp \hspace*{2em} Netherlands \hspace*{2em} You \hspace*{2em} 1992 \hspace*{2em} 52 \hspace*{2em} F \\
2 \hspace*{2em} Timbaland \hspace*{2em} United States \hspace*{2em} Dangerous \hspace*{2em} 2008 \hspace*{2em} 32 \hspace*{2em} T \\
3 \hspace*{2em} Justin Brown \hspace*{2em} France \hspace*{2em} Hey Oh \hspace*{2em} 2013 \hspace*{2em} 29 \hspace*{2em} T \\
**/ \\
create table concert ( \\
\hspace*{2em} concert\_ID number, \\
\hspace*{2em} concert\_Name text, \\
\hspace*{2em} Theme text, \\
\hspace*{2em} Stadium\_ID text, \\
\hspace*{2em} Year text, \\
\hspace*{2em} primary key (concert\_ID), \\
\hspace*{2em} foreign key (Stadium\_ID) references stadium(Stadium\_ID) \\
) \\
/* \\
3 example rows from table concert: \\
concert\_ID \hspace*{2em} concert\_Name \hspace*{2em} Theme \hspace*{2em} Stadium\_ID \hspace*{2em} Year \\
1 \hspace*{2em} Auditions \hspace*{2em} Free choice \hspace*{2em} 1 \hspace*{2em} 2014 \\
2 \hspace*{2em} Super bootcamp \hspace*{2em} Free choice 2 \hspace*{2em} 2 \hspace*{2em} 2014 \\
3 \hspace*{2em} Home Visits \hspace*{2em} Bleeding Love \hspace*{2em} 2 \hspace*{2em} 2015 \\
**/ \\
create table singer\_in\_concert ( \\
\hspace*{2em} concert\_ID number, \\
\hspace*{2em} Singer\_ID text, \\
\hspace*{2em} primary key (concert\_ID), \\
\hspace*{2em} foreign key (Singer\_ID) references singer(Singer\_ID), \\
\hspace*{2em} foreign key (concert\_ID) references concert(concert\_ID) \\
) \\
/* \\
3 example rows from table singer\_in\_concert: \\
concert\_ID \hspace*{2em} Singer\_ID \\
1 \hspace*{2em} 2 \\
1 \hspace*{2em} 3 \\
1 \hspace*{2em} 5 \\
**/ \\
Question: How many singers do we have?

\end{document}